%% file: 3831.tex
\documentclass[runningheads]{llncs}
\pdfoutput=1
\usepackage{epsfig}
\usepackage{graphicx}
\usepackage{amsmath}
\usepackage{amssymb}

\usepackage[utf8]{inputenc} 
\usepackage[T1]{fontenc}    
\usepackage{url}            
\usepackage{booktabs}       
\usepackage{amsfonts}       
\usepackage{nicefrac}       
\usepackage{microtype}      

\usepackage{graphicx}
\usepackage{amsmath,amssymb} 
\usepackage{bbold}
\usepackage{dsfont}
\usepackage{enumerate}
\usepackage{enumitem}
\usepackage[toc,page]{appendix}

\usepackage{xcolor}
\usepackage{bm}
\usepackage{isomath}
\usepackage{arydshln}
\usepackage{stmaryrd}

\usepackage{pbox}
\usepackage{capt-of}

\usepackage[normalem]{ulem}
\usepackage{multirow}

\usepackage{colortbl}
\usepackage{xspace}
\usepackage{mathtools}
\usepackage{array}
\usepackage{gensymb}
\usepackage{ifpdf}
\usepackage{refcount}
\usepackage[toc,page]{appendix}

\makeatletter
\DeclareRobustCommand\bmvaOneDot{\futurelet\@let@token\bmv@onedotaux}
\def\bmv@onedotaux{\ifx\@let@token.\else.\null\fi\xspace}
\makeatother

\def\etal{\emph{et al}\bmvaOneDot}
\def\etc{\emph{etc}\bmvaOneDot}
\def\ie{\emph{ie}\bmvaOneDot}
\def\wrt{w.r.t\bmvaOneDot}
\def\vs{\emph{vs}\bmvaOneDot}

\newcommand{\comment}[1]{}

\usepackage{indentfirst}
\input{definitions.tex}

\begin{document}

\pagestyle{headings}
\mainmatter
\def\ECCVSubNumber{3831}  

\title{Few-shot Action Recognition with Permutation-invariant Attention} 

\titlerunning{Few-shot AR with Permutation-invariant Attention}
%
\author{Hongguang Zhang\inst{1,2,3,5}, 
Li Zhang\inst{2}, 
Xiaojuan Qi\inst{4,2}, 
Hongdong Li\inst{1,5}, \\
Philip H. S. Torr\inst{2}, 
Piotr Koniusz\inst{3,1}}
\authorrunning{H. Zhang \etal}
%

\institute{Australian National University, Canberra, Australia \and
University of Oxford, Oxford, UK \and
Data61/CSIRO, Australia \and
The University of Hong Kong, Hong Kong, China \and
Australian Centre for Robotic Vision, Australia}
\maketitle

\setcounter{footnote}{1}

\input{0-abstract.tex}
\input{1-introduction.tex}
\input{2-relatedwork.tex}
\input{3-method.tex}

\input{4-experiments.tex}

\input{5-conclusion.tex}

{
\vspace{0.1cm}
\noindent
\textbf{Acknowledgements.}
This research is supported in part by the Australian Research Council through Australian Centre for Robotic Vision (CE140100016), Australian Research Council grants (DE140100180), the China Scholarship Council (CSC Student ID 201603170283). Hongdong Li is  funded in part by ARC-DP (190102261) and ARC-LE (190100080). We  thank  CSIRO Scientific Computing, NVIDIA (GPU grant) and the National University of Defense Technology. 
}

\section*{Appendix}
\input{appendix.tex}

\input{3831.bbl}
\end{document}

%% file: definitions.tex
\renewcommand\vec[1]{\ensuremath\boldsymbol{#1}}
\renewcommand\cdots{...}

\newcommand{\tD}{\vec{\mathcal{D}}}

\newcommand{\tF}{\vec{\mathcal{F}}}
\newcommand{\tT}{\vec{\mathcal{T}}}

\newcommand{\mX}{\mathbf{X}}

\newcommand{\mbr}[1]{\mathbb{R}^{#1}}

\newcommand{\tR}{\vec{\mathcal{R}}}

\newcommand{\vphi}{\boldsymbol{\phi}}
\newcommand{\vpsi}{\boldsymbol{\psi}}

\newcommand{\mPsi}{\vec{\Psi}}

\DeclareMathOperator*{\argmin}{arg\,min}

\def\eg{\emph{e.g.}}

\newcommand{\vPhi}{\boldsymbol{\Phi}}

\newcommand{\tS}{\vec{\mathcal{S}}}

\newcommand{\mPhi}{\boldsymbol{\Phi}}

\newcommand{\stkout}[1]{{\ifmmode\text{\sout{\ensuremath{#1}}}\else\sout{#1}\fi}}

%% file: 0-abstract.tex
\begin{abstract}
Many few-shot learning models focus on recognising images.  
In contrast, we tackle a challenging task of few-shot action recognition from videos.  We build on a C3D encoder for spatio-temporal video blocks to capture short-range action patterns. Such encoded blocks are aggregated by permutation-invariant pooling to make our approach robust to varying action lengths and long-range temporal dependencies whose patterns are unlikely to repeat even in clips of the same class. Subsequently, the pooled representations are combined into simple relation descriptors which encode so-called query and support clips. Finally, relation descriptors are fed to the comparator with the goal of similarity learning between query and support clips. Importantly, to re-weight block contributions during pooling, we exploit spatial and temporal attention modules and self-supervision. In naturalistic clips (of the same class) there exist a temporal distribution shift--the locations of discriminative temporal action hotspots vary. Thus, we permute blocks of a clip and align the resulting attention regions with similarly permuted attention regions of non-permuted clip to train the attention mechanism invariant to block (and thus long-term hotspot) permutations. Our method outperforms the state of the art on the HMDB51, UCF101,  \textit{mini}MIT datasets.
\end{abstract}

%% file: 1-introduction.tex
\section{Introduction}

Few-shot learning is an open problem with the goal to design algorithms that learn in the low-sample regime. Examples include meta-learning  \cite{finn2017model,lee2019meta,antoniou2018train,Simon_2019_NIPS,Simon_2020_CVPR},  robust feature representations by relation learning \cite{snell2017prototypical,sung2017learning,wertheimer2019few,zhang2019power}, gradient-based \cite{Zintgraf2019Cavia,Rusu2019LEO,Simon_2020_ECCV} and  hallucination strategies for insufficient data \cite{hariharan2017low,zhang2019few}. 

\begin{figure}[t]
    \centering
    \includegraphics[width=\linewidth]{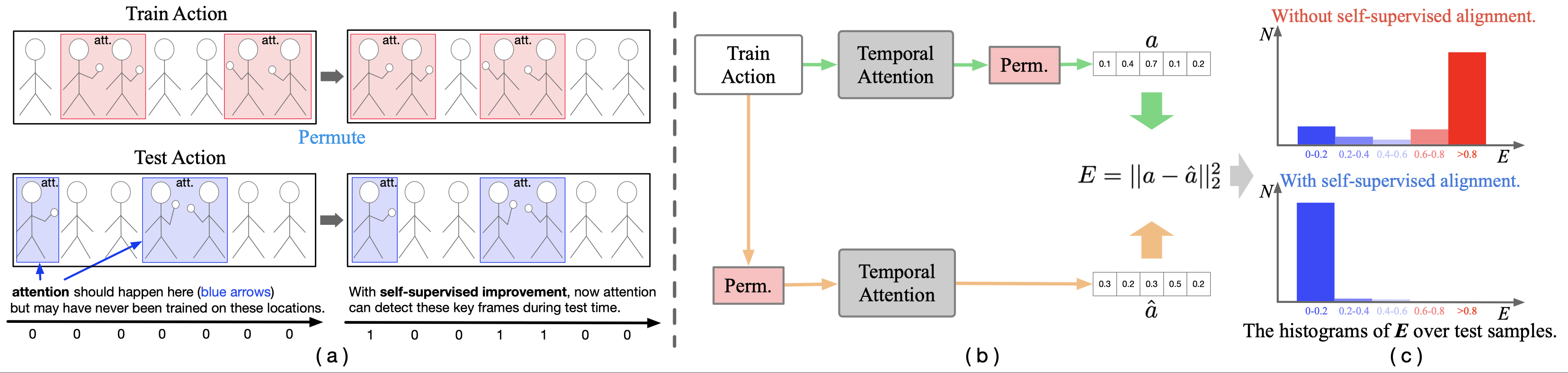}
    \caption{Augmentation-guided attention by alignment. Fig. \ref{fig:demo}a shows that discriminative action blocks (in red, top left) may be misaligned with discriminative action blocks of test clip (in blue, bottom left). If the attention unit observes different distributions of locations of discriminative blocks at training and testing time, it fails. With the right approach, one may overcome the distribution shift (top and bottom right panels). Fig. \ref{fig:demo}b shows how the augmentation-guided attention by alignment works for permutation-based augmentations: (i) we shuffle training  blocks of a clip to train an attention on permuted blocks, (ii) we shuffle in the same way coefficients of the attention vector from the non-permuted blocks. Both attention vectors are then encouraged to align by a dedicated loss term during training. Fig. \ref{fig:demo}c shows histograms of alignment-errors on test data. The top histogram shows larger errors (no alignment loss used in training) while the bottom histogram shows small errors (alignment during testing improves).
    }
    \label{fig:demo}
\end{figure}

However, very few papers address video-based few-shot learning.  As annotating large video datasets is prohibitive, this makes the problem we study particularly valuable.  While results are far from satisfactory on {\em Kinetics} \cite{i3d_net}, the largest action recognition dataset, its size of 300,000 video clips with hundreds of frames each exceeds the size of large-scale image 
ImageNet \cite{ILSVRC15} and Places205 \cite{places_dataset} datasets. 

There exist few limited works on few-shot learning for action recognition \cite{mishra2018generative,xu2018dense,guo2018neural,Zhu_2018_ECCV}. However, they focus on modeling 3D body joints with graphs \cite{guo2018neural}, attribute-based learning in generative models \cite{mishra2018generative},   network design for low-sample classification \cite{xu2018dense} and salient memory approach \cite{Zhu_2018_ECCV}. In contrast, we focus on robust  relation/similarity, spatial and temporal modeling of short- and long-term range action patterns via permutation-invariant pooling and attention. 

To obtain a robust few-shot action recognition approach, we investigate how to: (i) represent discriminative short- and non-repetitive long-term action patterns for relation/similarity learning, (ii) localize temporally  discriminative action blocks with limited number of training samples, and (iii) deal with long-term temporal distribution shift of such discriminative patterns (these patterns never re-appear at the same temporal locations even for clips of the same class).

\footnotetext{\label{foot:selfsup}Self-supervision assumes generating cheap-to-obtain data (\eg, augmentation by rotations) from the original data and imposing an auxiliary task whose goal is to predict the label of  an augmentation with the goal of robust representation learning \cite{dosovitskiy2014discriminative,doersch2015unsupervised,gidaris2018unsupervised,jian2019evolutionarily,jian2020representation}. We are first to apply the self-supervision by alignment paradigm to the problem of robust attention training (we devise an augmentation-guided attention).}

To address the first point, our early experiments indicated that short-term discriminative action patterns can be captured by an encoder with  C3D convolutional blocks. Thus, resulting features from a clip undergo permutation-invariant pooling which discards long-term non-repetitive dependencies. Finally, pooled query/support representations form relation descriptors are fed into a comparator.

Regarding the second point, aggregating  spatio-temporal blocks with equal weights is suboptimal. Thus, we devise spatial/temporal attention units to emphasize discriminative blocks. Under the low-sample regime, self-supervision by {\em jigsaw} and {\em rotation} \footnotemark[\getrefnumber{foot:selfsup}] helps train a more robust  encoder, comparator and attention. However, vanilla attention (and/or  self-supervision) cannot fully promote the invariance to temporal (or spatial) permutations as described next.

To address the third point, we note that long-term dependencies in clips are non-repetitive \eg, videos of the same class often contain relevant action blocks at different temporal locations. Figure \ref{fig:demo}a shows that discriminative blocks of training and testing clips of {\em dance} class do not align (top left \vs bottom left corner). By permuting the blocks of training (top right), one can make them align with the most discriminative test samples (bottom right). Figure \ref{fig:demo}b shows that for a given clip, we (i) shuffle its blocks and feed them to the attention mechanism (shuffling pass), (ii) we shuffle accordingly the attention coefficients from a non-shuffled pass through attention, and (iii) we force attention coefficients from both passes to align. Such an attention by alignment deals with the distribution shift of discriminative temporal (and spatial) patterns via {\em jigsaw} augmentation (but applies also to {\em rotation, zoom}, \etc) 
To summarize, our contributions include:
\renewcommand{\labelenumi}{\roman{enumi}.}
\begin{enumerate}[leftmargin=0.6cm]
    \item  A robust pipeline with a C3D-based encoder capturing short-term dependencies which yields block representations subsequently aggregated by permutation-invariant pooling into fixed-length representations which form relation descriptors for relational/similarity learning in an episodic setting \cite{sung2017learning}.
    \item Spatial and temporal attention units which re-weight block contributions during the aggregation step. To improve training of the encoder, comparator and the attention unit  under the low-sample regime, we introduce spatial and temporal self-supervision by {\em rotations}, and {\em spatial} and {\em temporal jigsaws}.
    \item  An improved self-supervised attention unit by applying augmentation patterns such as {\em jigsaws} and/or {\em rotations} on the input of the attention unit and aligning the output with augmented the same way attention vector coefficients from non-augmented data passed by the attention unit\footnotemark[\getrefnumber{foot:selfsup}]. Thus, the attention unit becomes invariant to a given augmentation action by design.
    \item We propose new data splits for a systematic comparison of few-shot action recognition algorithms and we make them available as existing approaches use each different pipeline concepts, data modality, data splits and  protocols\footnote{Section \ref{sec:contrast} explains that existing works do not specify class/validation splits which yields $\sim$6\% variations in accuracy rendering their protocols highly inaccurate. Section \ref{sec:issue} explains this issue  and how we compare our method to existing works.}.
\end{enumerate}


%% file: 2-relatedwork.tex
\section{Related work}
\label{sec:related}

Below, we discuss zero-, one- and few-shot learning models followed by a discussion on self-supervised learning and second-order pooling.



\noindent{\textbf{One- and few-shot learning}} models have been widely studied in both the shallow \cite{miller_one_example,Li9596,NIPS2004_2576,BartU05,fei2006one} and  deep learning pipelines \cite{koch2015siamese,vinyals2016matching,snell2017prototypical,finn2017model,snell2017prototypical,sung2017learning}. 
Motivated by the human ability to learn new concepts from few samples, early works \cite{fei2006one,lake_oneshot} employ generative models with an iterative inference. 
Siamese Network \cite{koch2015siamese} is a two-stream convolutional neural network  which generates image descriptors and learns the similarity between them. 
Matching Network \cite{vinyals2016matching}  proposes query-support episodic learning and $L$-way $Z$-shot learning protocols\footnote{Kindly see \cite{vinyals2016matching,snell2017prototypical,sung2017learning} for the concept of query, support and episodic learning, and the evaluation protocols which  differ from traditional recognition and low-shot learning.}. The similarity between a query and support images is learnt for each episode. At the testing time, each test query (of novel class) is compared against a handful of annotated test support images for rapid recognition. 
%
%
Prototypical Networks \cite{snell2017prototypical}  compute distances between a datapoint and class-wise prototypes. %
Model-Agnostic Meta-Learning (MAML) \cite{finn2017model} 
is trained on multiple  learning tasks. 
Relation Net \cite{sung2017learning} 
learns relations between query and support images, 
and it leverages a similarity learning neural network to compare query-support pairs. 
SalNet \cite{zhang2019few} uses saliency-guided end-to-end sample hallucination to grow the training set. 
Graph Neural Networks (GNN) have also been used in few-shot learning \cite{gnn,Kim_2019_CVPR,Gidaris_2019_CVPR}. 

\noindent{\textbf{Self-supervised learning}} leverages 
free supervision signals residing in images and videos
to promote robust representation learning in image recognition  \cite{dosovitskiy2014discriminative,doersch2015unsupervised,gidaris2018unsupervised}, video recognition \cite{fernando2017self,sermanet2017time,gan2018geometry}, video object segmentation~\cite{lai2020mast,zhu2020self}
and few-shot image classification \cite{gidaris2019boosting,su2019boosting}. 
Approaches \cite{gidaris2018unsupervised,doersch2015unsupervised,dosovitskiy2014discriminative} learn to predict random image rotations, relative pixel positions, and surrogate classes, respectively. Finally, \cite{gidaris2019boosting,su2019boosting} improve few-shot results by predicting image rotations/jigsaw patterns. 

\noindent{\textbf{Second-order statistics}} are used by us for permutation-invariant pooling. They
are also used for texture recognition \cite{tuzel_rc,elbcm_brod} by so-called Region Covariance Descriptors (RCD), object and action recognition \cite{koniusz2017higher,lei_beycov,pk_tpami_ar}. 
Second-order pooling has also been used in fine-grained image classification \cite{koniusz2018deeper,DeepKSPD,lei_beycov}, domain adaptation \cite{koniusz2017domain} and the fine-grained few-shot learning \cite{zhang2019power,wertheimer2019few,pk_hz_tpami_fsl}.

\noindent{\textbf{Few-shot action recognition}} 
approaches \cite{mishra2018generative,guo2018neural,xu2018dense} use a generative model, graph matching on 3D coordinates and a dilated networks with class-wise classifiers, respectively. Approach \cite{Zhu_2018_ECCV} proposes a so-called compound memory network using key-value memory associations. 
ProtoGAN \cite{dwivedi2019protogan} proposes a GAN model to generate action prototypes to address few-shot action recognition.

\subsection{Contrast with existing works}
\label{sec:contrast}
%
Unlike  \cite{guo2018neural}, we use video clips rather than 3D skeletal coordinates. In contrast to \cite{xu2018dense}, we use relation/similarity learning and our training/testing class concepts are disjoint. While \cite{Zhu_2018_ECCV} memorizes  key values/frames, we model short- and long-term dependencies. 
While \cite{dwivedi2019protogan} forms action prototypes by GAN, we focus on self-supervised attention learning and permutation-invariant aggregation.

In contrast to self-supervision by rotations and jigsaw \cite{gidaris2019boosting,su2019boosting}, we use a sophisticated self-supervision on the attention unit for which a dedicated loss performs the alignment between the attention vector of augmented attention unit and the augmented in the same way attention vector from the non-augmented attention unit. Thus, we train a permutation-invariant attention to deal with the distribution shift of discriminative action locations.

Finally, we use second-order pooling \cite{koniusz2018deeper,zhang2019power}  for a permutation-invariant aggregation of temporal blocks while \cite{koniusz2018deeper,zhang2019power} work with images. We develop a theory explaining why Power Normalization helps episodic learning.

\subsection{Issues with fair comparisons}
\label{sec:issue}
Each few-shot action recognition method from Section \ref{sec:related} uses  different datasets and evaluation protocols making fair comparisons impossible. Class-wise splits/vali- dation sets are unavailable \ie, model \cite{mishra2018generative} uses a random split. Figures \ref{fig:ssta_loss}c and \ref{fig:ssta_loss}d of Section \ref{sec:exp} show that the random choice of the split set yields up to $\sim$6\% deviation in accuracy rendering such a protocol problematic. 
Thus, we propose a new protocol with class splits and validation sets made publicly available.

%% file: 3-method.tex
\section{Approach}

\label{sec:approach}

\begin{figure*}[t]
    \centering
    \includegraphics[width=\linewidth]{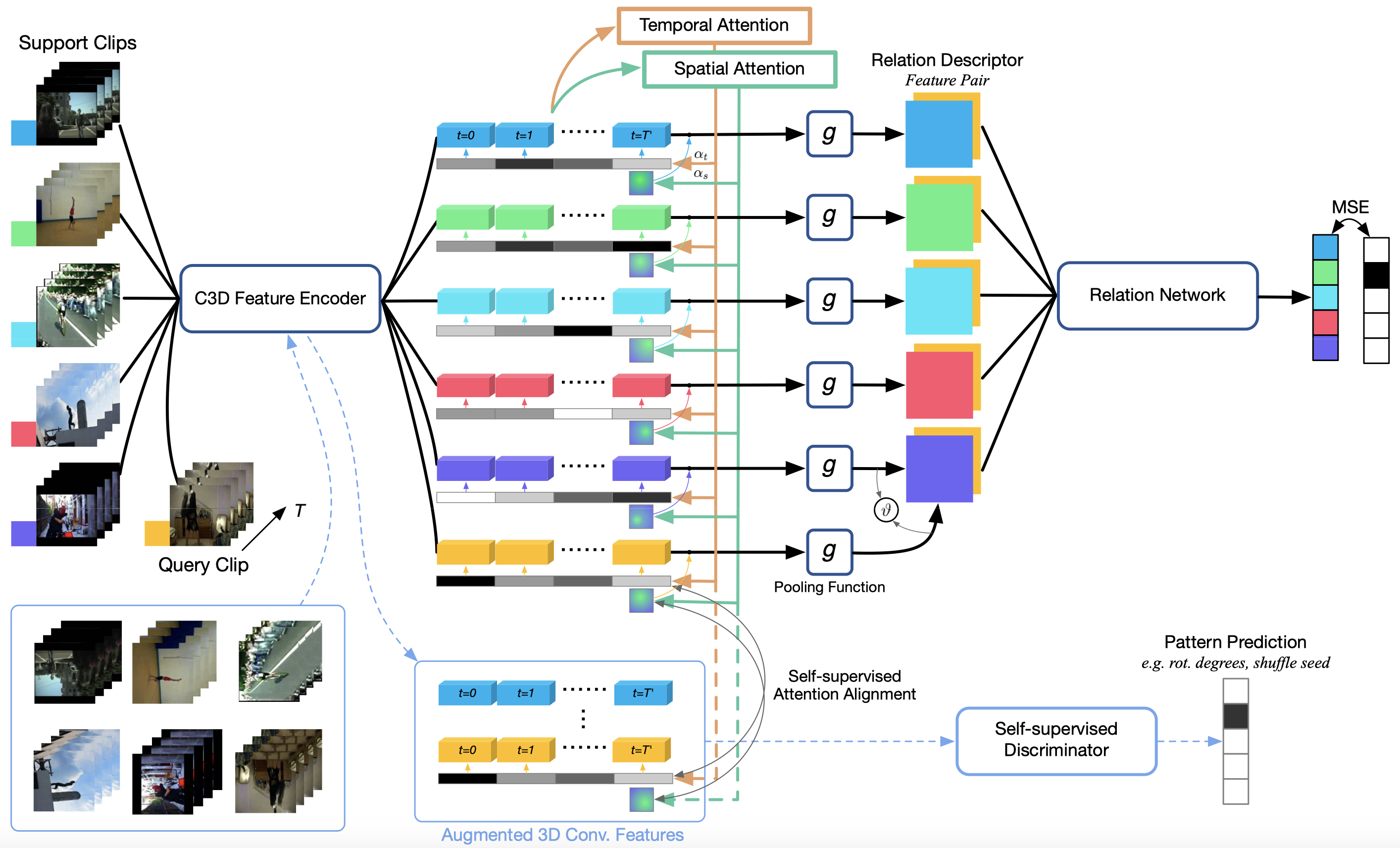}
    \caption{\small Our few-shot Action Relation Network (ARN) contains: feature encoder with 4-layer 3D conv. blocks, relation network with 2D conv. blocks, and spatial and temporal attention units which refine the aggregation step. Specifically, we apply second-order pooling (operator $g$) over encoder outputs (re-weighted by attention vectors) per clip to obtain a Power Normalized Autocorrelation Matrix (AM). Query and support AMs per episode form relation descriptors (by operator $\vartheta$) from which the relation network learns to capture relations. The block (blue dashed line) is the self-supervised learning module which encourages our pipeline to learn auxiliary tasks \eg, {\em jigsaws}, {\em rotations}.}
    \label{fig:pipe}
\end{figure*}

\subsection{Pipeline}
Figure \ref{fig:pipe} shows our Action Relation Network (ARN). In contrast to the \textit{Conv-4-64} backbone in few-shot image classification  \cite{sung2017learning,zhang2019power,zhang2019few}, we adopt a C3D-based \textit{Conv-4-64} backbone to extract spatio-temporal features which capture short-range dependencies. Next, we apply second-order pooling on 3D action features re-weighted by attention to obtain second-order statistics which are permutation-invariant \cite{koniusz2018deeper} \wrt the spatio-temporal order of features. To paraphrase, we discard the long-range order of temporal (and spatial) blocks captured by the encoder. Finally, second-order matrices form relation descriptors from query/support clips fed into a 2D relation network to capture relations. 

Let $\mathbf{V}$ denote a video (\ie, with $\sim\!$20 frames) and $\mPhi\!\in\!\mbr{C\times T\times H\times W}$ be features extracted from $\mathbf{V}$  by $f$:
\begin{equation}
    \mPhi = f(\mathbf{V}; \tF).
    \label{eq:enc}
\end{equation}

\begin{figure}[t]
    \centering
    \includegraphics[width=\linewidth]{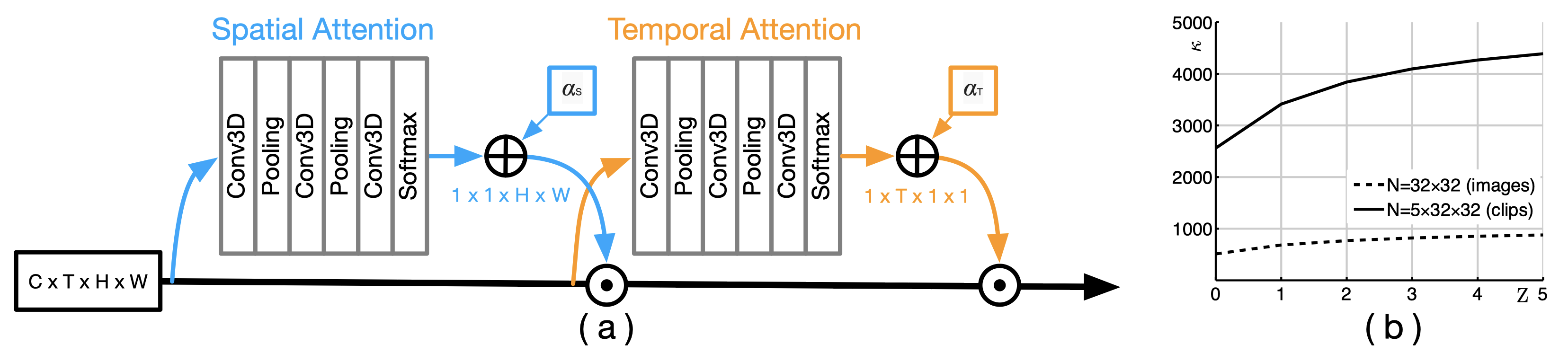}
    \caption{\small Spatial and temporal units are shown in Fig. \ref{fig:ar_attention}a. A naive approach is to directly extract the temporal and spatial attention, whose size is $1\times T\times H\times W$. However, this is computationally expensive and results in overparametrization. Thus, we split the attention block into separate spatial and temporal branches whose impact is adjusted by $\alpha_s$ and $\alpha_t$. Fig. \ref{fig:ar_attention}b is the $\kappa$ ratio \wrt~the $Z$-shot value (see Eq. \eqref{eq:support}). The dashed curve shows that as $Z$ grows (0 denotes the regular classification), the memorization burden of co-occurrence $(i,j)$ on the comparator grows $\kappa$ times for second-order pooling without Power Normalization (as opposed to Power Normalization). The solid line shows that as we use larger $N$ (video clips \vs images), not using PN is even more detrimental.}
    \label{fig:ar_attention}
\end{figure}


To aggregate $\mPhi$ per clip into $\mPsi$, we apply a pooling operator $g$ over the support and query features, resp. For $g$, we use pooling operators from Sec. \ref{sec:pool}:  

\begin{equation}
    \mPsi = g(\mPhi).
\end{equation}

Once $\mPsi$ are computed for query/support clips, they form relation descriptors (via operator $\vartheta$) passed to the relation network $r$ to obtain the relation score $\zeta_{sq}$:
\begin{equation}
    \zeta_{sq} = r(\vartheta(\mPsi_s, \mPsi_q); \tR),
\end{equation}
where $\tR$ are parameters of network $r$, and $\vartheta$ forms relation descriptors \eg, we use the concatenation along the channel mode.

We use the Mean Square Error (MSE) loss over support and query pairs:
\begin{equation}
    L = \sum\limits_{s\in S}\sum\limits_{q\in Q} (\zeta_{sq} - \delta(l_s-l_q))^2,
\end{equation}
where $\delta(l_s\!-\!l_q)\!=\!1$ if $l_s\!=\!l_q$, $\delta(l_s\!-\!l_q)\!=\!0$ otherwise. Class labels of support and query action clips are denoted as  $l_s$ and $l_q$.

\begin{figure*}[t]
    \centering
    \includegraphics[width=\linewidth]{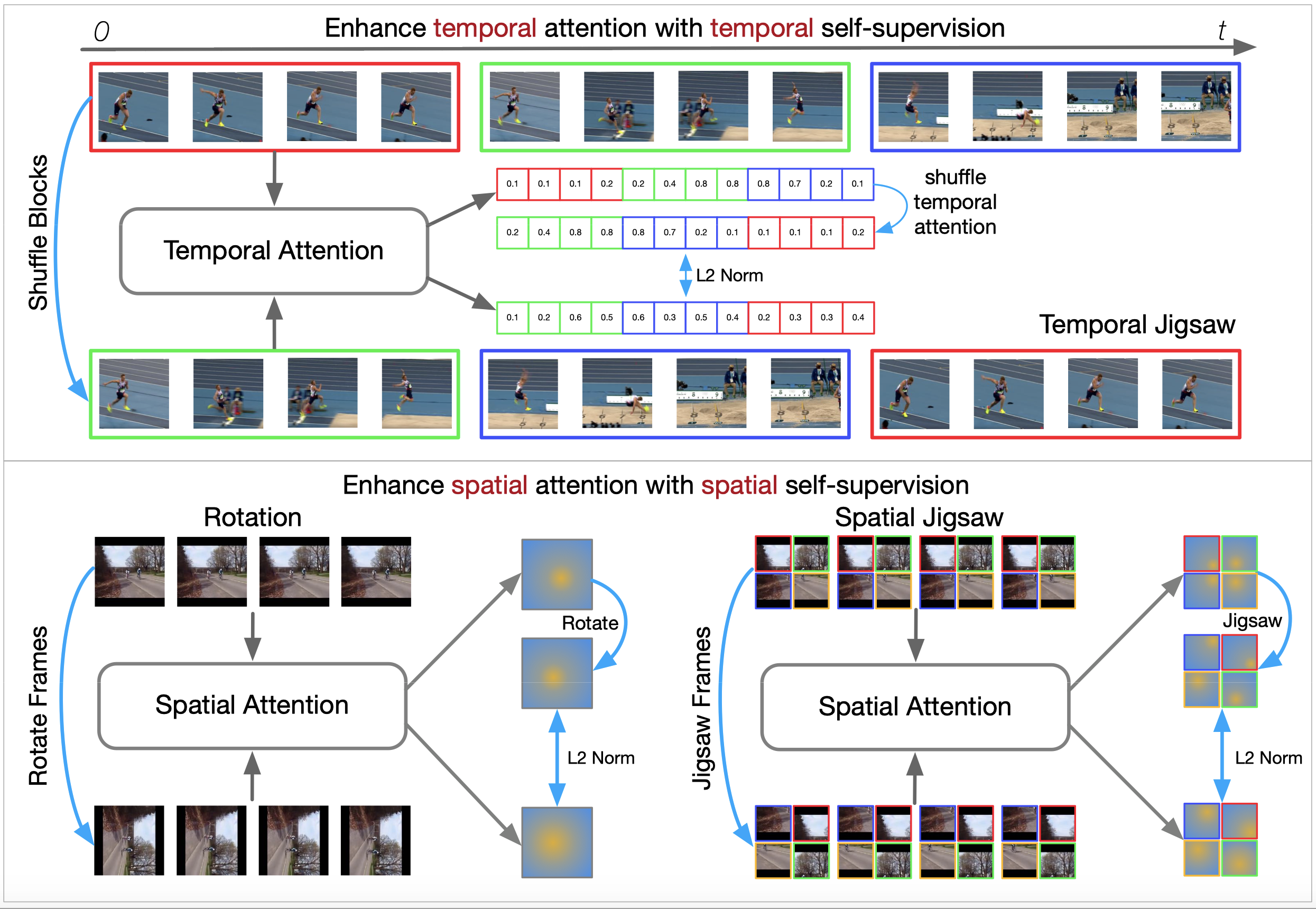}
    \caption{Augmentation-guided attention by alignment. We firstly collect the encoded representations of original and augmented data, then we extract the temporal or spatial attention vectors from them. We apply the same augmentation(s) on the temporal or spatial attention vectors of the original data resulting in the augmented attention vectors which we align with attention vectors of the augmented data.}
    \label{fig:ss_attention}
\end{figure*}

\subsection{Pooling of encoded representations}
\label{sec:pool}
For permutation-invariant pooling of temporal (and spatial) blocks, we investigate three pooling mechanisms discussed below.

\noindent{\textbf{Average and max pooling}} are two widely-used pooling functions which can be used for aggregation of $N\!=\!T\!\times\!W\!\times\!H$ fibers (channel-wise vectors) of feature map $\mPhi$ defined in Eq. \eqref{eq:enc}. The average pooling is given as $\vpsi\!=\! \frac{1}{N}\sum_{n=1}^N\vphi_n$ where $\vpsi\!\in\!\mbr{C}$, and $\vphi_n\!\in\!\mbr{C}$ are $N$ fibers. Similarly, max pooling is given by $\psi_c\!=\!\!\! \max\limits_{n=1,\cdots,N}\phi_{cn},\, c\!=\!1,\cdots,C$, and $\vpsi=[\psi_1,\cdots,\psi_C]^T$. Average and max pooling are commutative \wrt~the input fibers, thus being permutation-invariant. However, first-order pooling is less informative than second-order \cite{koniusz2016tensor} discussed next.

\noindent{\textbf{Second-order pooling}} captures correlations (or co-occurrences) between pairs of features in $N$ fibres of feature map $\mPhi$, which is reshaped such that $\mPhi\in\mbr{C\times N}$, $N=T\times H\times W$. Such an operator proved robust in classification \cite{koniusz2016tensor} and  few-shot learning \cite{zhang2019power,wertheimer2019few,pk_hz_tpami_fsl}. 
Specifically, we define:
\begin{align}
& \!\!\!\mPsi=\eta\Big(\frac{1}{N}\sum_{n=1}^N\!\vphi_n\vphi_n^T\Big)=\eta\Big(\frac{1}{N}\mPhi\mPhi^T\Big)\quad\text{where}\quad\eta(\mX) = \frac{1 - \exp({\sigma \mX})}{1 + \exp({\sigma \mX})}.\label{eq:hok3}
\end{align}
%
%

\noindent{Matrix} $\mPsi\in\mbr{C\times C}$ is a Power Normalized autocorrelation matrix capturing correlations of fiber features $\vphi_n$ of feature map $\mPhi$ from Eq. \eqref{eq:enc} while $\eta$ applies Power Normalization (PN): we use the zero-centered element-wise Sigmoid \cite{koniusz2016tensor,zhang2019power} on $\mX$, and
%
 $\sigma$ controls the slope of PN. For a given pair of features $i$ and $j$ in matrix $\mPsi$, that is $\Psi_{ij}$, the role of PN is to  detect the likelihood if at least one co-occurrence of features $i$ and $j$ has been detected \cite{koniusz2016tensor}.
%
According to Eq. \eqref{eq:hok3}, second-order pooling is permutation-invariant \wrt~the order of input fibers as the summation in Eq. \eqref{eq:hok3} is commutative \wrt~the order of $\vphi_1,\cdots,\vphi_N$. Thus, second-order pooling factors out the spatial and temporal modes of $\mPhi$ and aggregates clips with varying numbers of temporal blocks (discards the order of long-range spatial/temporal dependencies) into a fixed length representation $\mPsi\in\mbr{C\times C}$. Below we explain further why second-order pooling with Power Normalization is  well suited for episodic few-shot learning.

\noindent{\textbf{Relation descriptors}} between query/support pooled matrices $\mPsi_q$ and $\mPsi_s$ are formed by operation $\vartheta(\mPsi_q,\mPsi_s)$ which, in our case, simply performs concatenation of $\mPsi_q$ with $\mPsi_s$ along the channel mode by $\text{cat}(\mPsi_q, \mPsi_s)\in\mbr{2\times C\times C}$, and $\mPsi_s$ is obtained by the mean (or maximum) along the channel mode between $\mPsi_s^{1},\cdots,\mPsi_s^{Z}$  belonging to the same episode and class ($Z\!>\!1$ for the few-shot case).

It is known from \cite{koniusz2018deeper} that the Power Normalization in Eq. \eqref{eq:hok3} performs a co-occurrence detection rather than counting (correlation). For classification problems, assume a probability mass function $p_{X_{ij}}(x)\!=\!1/(N\!+\!1)$ if $x\!=\!0,\cdots,N$, $p_{X_{ij}}(x)\!=\!0$ otherwise, that tells the probability that co-occurrence between $\Phi_{in}$ and $\Phi_{jn}$ happened $x\!=\!0,\cdots,N$ times (given some clip). Note that  classification often  depends on detecting a co-occurrence (\eg, is there a flower co-occurring with a pot?) rather than counts (\eg, how many flowers and pots co-occur?). Using second-order pooling without PN requires a classifier to observe $N\!+\!1$ training samples of {\em flower and pot} co-occurring in quantities $0,\cdots,N$ to memorise all possible co-occurrence count configurations. For relation learning, our $\vartheta$ stacks pairs of samples to compare, thus a comparator now has to deal with a probability mass function of ${R_{ij}}\!=\!{X_{ij}}\!+\!{Y_{ij}}$ depicting {\em flowers and pots} whose  $\text{support}(p_{R_{ij}})\!=\!2N\!+\!1\!>\!\text{support}(p_{X_{ij}})\!=\!N\!+\!1$ if random variable $X\!=\!Y$ (same class). The same is shown by variances  $\text{var}(p_{R_{ij}})\!>\!\text{var}(p_{X_{ij}})$. For $Z$-shot learning, the growth of variance and support equal $(Z\!+\!1)N\!+\!1$  indicates that the comparator has to memorize more configurations of co-occurrence $(i,j)$ as $Z$ grows.

However, this situation is alleviated by Power Normalization (operator $\eta$) whose  probability mass function can be modeled as $p_{X^\eta_{ij}}(x)\!=\!1/2$ if $x\!=\!\{0,1\}$, $p_{X^\eta_{ij}}(x)\!=\!0$ otherwise, as PN detects a co-occurrence (or its lack). For $Z$-shot learning, $\text{support}(p_{R^\eta_{ij}})\!=\!Z\!+\!2\!\ll\!\text{support}(p_{R_{ij}})\!=\!(Z\!+\!1)N\!+\!1$. The  ratio given as
\begin{align}
\kappa\!=\!\frac{\text{support}(p_{R_{ij}})}{\text{support}(p_{R^\eta_{ij}})}\!=\!\frac{(Z\!+\!1)N\!+\!1}{Z\!+\!2}
\label{eq:support}
\end{align}

\noindent{shows} that the comparator has to memorize many more count configurations of co-occurrence $(i,j)$ for naive pooling compared to PN as $Z$ and/or $N$ grow ($N$ depends on the number of temporal and spatial blocks $T$, $H$ and $W$). Figure \ref{fig:ar_attention} shows how $\kappa$ varies \wrt~$Z$ and $N$. Our modeling assumptions are simple \eg, the assumption on mass functions with uniform probabilities, the use of the support of mass functions rather than variances to describe variability of co-occurrence $(i,j)$. Yet, substituting these modeling choices with more sophisticated ones does not affect  theoretical conclusions that: (i) PN benefits few-shot learning ($Z\!\geq\!1$) more than the regular classification ($Z\!=\!0$) in terms of reducing possible count configurations of $(i,j)$, and (ii) for videos (large $N$) PN reduces the number of count configurations of $(i,j)$ more rapidly than for images (smaller $N$). While classifiers and comparators do not learn exhaustively all count configurations of co-occurrence $(i,j)$ as they have some generalization ability, they learn quicker and better if the number of count configurations of $(i,j)$ is limited. 

\subsection{Temporal and spatial attention}
Figure \ref{fig:ar_attention}a introduces decoupled spatial/temporal attention units consisting of three 3D Convolutional blocks and a Sigmoid output layer. Let $t$ and $s$ denote the temporal and spatial attention modules, and  the attention be applied ahead of second-order pooling. We obtain temporal and spatial attention maps  $\mathbf{T}\in\mbr{1\times T\times 1\times 1}$ and  $\mathbf{S}\in\mbr{1\times 1\times \times H\times W}$, and attentive action features $\mPhi^*$ by:
\begin{align}
    \mathbf{T} = t(\mPhi; \tT), \;\mathbf{S} = s(\mPhi; \tS), \\
    \mPhi^* = (\alpha_t + \mathbf{T})\cdot(\alpha_s +\mathbf{S})\cdot\mPhi,\label{eq:sh}
\end{align}
where $\tT$ and $\tS$ are network parameters of temporal/spatial attention units while $\alpha_t$ and $\alpha_s$ control the impact of attention vectors.

Using attention helps spot discriminative temporal/spatial blocks, and suppress uninformative regions. However, the attention  should be robust to varying distributions of locations of discriminative blocks in clips as proposed below.

\subsection{Temporal and spatial self-supervision}
Self-supervised Learning (SsL) helps 
learn representations without using manually-labeled annotations. 
We  impose self-supervision both on encoders and attention units. 
%
For temporal self-supervision, we augment clips by shuffling the order of temporal blocks, which primes our network to become robust to long-term non-repetitive temporal dependencies in clips. Self-supervision also helps overcome the low-sample 
by encouraging network to learn  auxiliary tasks. In contrast, previous works 
shuffled frames which breaks the highly discriminative short-term temporal dependencies. 
We use the following self-supervision strategies:
%
\renewcommand{\labelenumi}{\roman{enumi}.}
\begin{enumerate}[leftmargin=0.6cm]
    \item \textbf{Temporal jigsaw.} Jigsaw, a popular self-supervisory task breaks the object location bias and teaches the network to recognize shuffling. As in \cite{Xu_2019_CVPR}, we split clips into non-overlapping fixed-length temporal blocks and shuffle them. 
    \item \textbf{Spatial jigsaw.} We split frames into four non-overlapping regions, then randomly permute them.
    \item \textbf{Rotation.} As the most popular self-supervisory task are rotations, we uniformly rotate all frames per clip by a random angle ($0^{\circ}$, $90^{\circ}$, $180^{\circ}$, $270^{\circ}$).
\end{enumerate}

Figure \ref{fig:pipe} (blue frame) shows how we apply and recognize the self-supervision patterns \eg, shuffling and rotation angles. 
Below, we illustrate self-supervision via  rotations. Consider the objective function $L_{rot}$ for self-supervised learning with a self-supervision discriminator $d$, where $\tD$ are parameters of $d$. Thus:
\begin{align}
   & \hat{\mPhi}_i = f(\;rot(\mathbf{V}_i, \theta)\; ;\tF), \\
&\mathbf{p}_{rot_i} = d(\hat{\mPhi_i}; \tD),\\
%
   & \!\!\!\!\!\!\!\!\!\!\!\!\!L_{rot} = - \sum\limits_{i} log\left(\frac{\exp({\mathbf{p}_{rot_{i}}[l_{\theta i}]})}{\sum_{s}\exp({\mathbf{p}_{rot_{i}}\![l_{s}]})}\right),
    \label{eq:ssrot}
\end{align}
%
%
where $\mathbf{V}_i$ is a randomly sampled clip, $\theta\in\{0\degree, 90\degree, 180\degree, 270\degree\}$ is a randomly selected rotation angle of a frame, $l_{\theta_i}\!\in\!\{0,1,2,3\}$ is the rot. label for sample $i$.

Combining the original loss function $L$ with such a self-supervision term $L_{rot}$ results in a self-supervised few-shot action recognition pipeline. However, this objective does not make the attention to be invariant to augmentations per se.

\subsection{Augmentation-guided attention by alignment}
Figure \ref{fig:ss_attention} presents a strategy in which we extract the attention vector for an augmented  clip, then we apply  the same augmentation to the attention vector obtained from the original non-augmented clip, and we encourage such a pair of augmentation vectors to align by a dedicated MSE loss. This encourages the attention unit to be invariant \wrt~a given augmentation type. Fig. \ref{fig:demo}a explains why the temporal permutation strategy  benefits few-shot learning while Figure \ref{fig:ss_attention} shows how to apply permutations and rotations.
As an example, for a rotation-guided spatial-attention we define the alignment loss $L_{att}$:
\begin{align}
\hat{\vPhi_i} & = f(\;rot(\mathbf{V}_i, \theta)\; ;\tF),  \\
S_i & = s(\vPhi_i; \tS), \; \hat{S}_i  = s(\hat{\vPhi}_i; \tS), \\
L_{att} & = \;\sum\limits_{i}\;||\;|\;rot(S_i, \theta) - \hat{S}_i\;| - \lambda \;||^2_F. 
\end{align}
\noindent{where} $\lambda$  controls the strictness of alignment. The final objective then becomes:
\begin{equation}
    \argmin\limits_{\tF, \tD, \tT, \tS}\quad L + \beta L_{ss} + \gamma L_{att}
    \label{eq:finloss}
\end{equation}
where $\beta$ and $\gamma$ are the hyper-parameters adjusted by cross-validation, $L_{ss}$ is a chosen type of self-supervision \eg, via rotations as introduced in Eq. \eqref{eq:ssrot}.

%% file: 4-experiments.tex
\section{Experiments}
\label{sec:exp}
\subsection{Experimental setup}
Below, we describe our setup and evaluations in detail. To exclude  complicated data pre-processing and frame sampling steps typically used in action recognition, we sample uniformly 20 frames along the temporal mode for each dataset. 

\begin{table*}[t]
\caption{Comparisons between our ARN model and existing works on HMDA51 and UCF101 splits proposed in \cite{mishra2018generative} and a Kinetics split from \cite{Zhu_2018_ECCV} (given 5-way acc.)}
\label{tab:compar}
\makebox[\textwidth]{
\setlength{\tabcolsep}{0.4em}
\fontsize{9}{10}\selectfont
\begin{tabular}{l|cccccc}
\toprule
\multirow{2}{*}{$\;\;\;\;\;\;$Model} & \multicolumn{2}{c}{HMDB51 \cite{mishra2018generative}} & \multicolumn{2}{c}{UCF101 \cite{mishra2018generative}} & \multicolumn{2}{c}{Kinetics \cite{dwivedi2019protogan}} \\
 & 1-shot & 5-shot & 1-shot & 5-shot & 1-shot & 5-shot \\ \hline
\textit{GenApp} $\;\;\;\;$\cite{mishra2018generative} & ${- }$ & ${52.5 \pm 3.10}$ & ${-}$ & ${78.6 \pm 2.1}$ & - & - \\
\textit{ProtoGAN} \cite{dwivedi2019protogan} & $34.7 \pm 9.20 $ & $54.0 \pm 3.90$ & $57.8 \pm 3.0$ & $80.2 \pm 1.3$ & - & - \\
\textit{CMN} $\;\;\;\;\;\;\;\;$\cite{Zhu_2018_ECCV} & - & - & - & - & 60.5 & 78.9 \\
\hline
\textit{Ours} & $\mathbf{44.6 \pm 0.9}$ & $\mathbf{59.1 \pm 0.8}$ & $\mathbf{62.1 \pm 1.0}$ & $\mathbf{84.8 \pm 0.8}$ & $\mathbf{63.7}$ & $\mathbf{82.4}$ \\
\bottomrule
\end{tabular}}
\end{table*}

\begin{table*}[t]
\centering
\caption{Ablations of different modules of ARN given our proposed HMDB51 protocol (given 5-way acc.) We used {\em spatial-jigsaw} for self-supervision.}
\label{tab:att}
\makebox[\textwidth]{
\setlength{\tabcolsep}{0.4em}
\fontsize{9}{7}\selectfont
\begin{tabular}{cccc|cc}
\toprule
Baseline & Spatial Attention & Self-supervision  & Alignment & 1-shot & 5-shot \\
\hline
\checkmark & & & & 40.83 & 55.18 \\
\checkmark & \checkmark & & & 41.27 & 56.12 \\
\checkmark & & \checkmark & & 44.19 & 58.50 \\
\checkmark & \checkmark & \checkmark & & 44.61 & 59.71 \\
\checkmark & \checkmark &  & \checkmark & 43.11 & 57.35 \\
\checkmark & \checkmark & \checkmark & \checkmark & \textbf{45.17} & \textbf{60.56}  \\
\bottomrule
\end{tabular}}
\end{table*}

\vspace{0.05cm}
\noindent\textbf{HMDB51} \cite{Kuehne11}  contains 6849 clips divided into 51 action categories, each with at least 101 clips, 
31, 10 and 10 classes selected for training,  validation and  testing.

\noindent\textbf{Mini Moments in Time (miniMIT)} \cite{monfortmoments} 
contains 200 classes and 550 videos per class. We select 120, 40 and 40 classes for training, validation and testing.

\noindent\textbf{UCF101} \cite{soomro2012ucf101}, action videos from Youtube, has 13320 video clips and 101 action classes. We randomly select 70 training, 10 validation and 21 testing classes. 

\noindent\textbf{Kinetics}, used by  \cite{Zhu_2018_ECCV} to select a subset  for few-shot learning, consists of  64, 12 and 24 training, validation and testing classes. We use it for comparisons.

Training, validation and testing splits on the first three datasets are detailed in our supplementary material while authors of \cite{Zhu_2018_ECCV} provide the split on Kinetics. 
The frames of action clips from all  datasets are resized to $128\times128$. All models are trained on training splits. Validation set is only used for cross-validation. 

\begin{table*}[t]
\centering
\caption{Evaluations on HMDB51 (5-way acc.) Attention: Temporal (TA), Spatial (SA). Self-super.: Temp. (TS), Spat. (SS), Self-Super. \& Alignment: Temp. (TSA), Spat. (SSA).$\!\!\!\!$}
\label{tab:1}
\makebox[\textwidth]{
\setlength{\tabcolsep}{1.2em}
\fontsize{10}{12}\selectfont
\begin{tabular}{lcc}
\toprule
Model & 1-shot & 5-shot  \\
\hline
& \multicolumn{2}{c}{Baseline} \\
\hline
\textit{C3D Prototypical Net} \cite{snell2017prototypical} & ${38.05 \pm 0.89}$ & ${53.15 \pm 0.90}$\\
\textit{C3D RelationNet} \cite{sung2017learning}  & ${38.23 \pm 0.97}$ & ${53.17 \pm 0.86}$\\
\textit{C3D SoSN} \cite{zhang2019power}  & ${40.83 \pm 0.96}$ & ${55.18 \pm 0.86}$\\
\hline
 \multicolumn{3}{c}{ \bf Temporal/Spatial Attention only (TA \vs~SA)} \\
\hline
\textit{ARN+TA}  & ${41.97 \pm 0.97}$ & ${57.67 \pm 0.88}$\\ 
\textit{ARN+SA}  & ${41.27 \pm 0.98}$ & ${56.12 \pm 0.89}$\\
\textit{ARN+SA+TA}  & ${42.41 \pm 0.99}$ & ${56.81 \pm 0.87}$\\
\hline
\multicolumn{3}{c}{\bf Temporal/Spatial Self-supervision only (TS \vs~SS)} \\
\hline
\textit{ARN+TS (temp. jigsaw)}  & ${43.79 \pm 0.96}$ & ${58.13 \pm 0.88}$\\
\textit{ARN+SS (spat. jigsaw)}  & ${44.19 \pm 0.96}$ & ${58.50 \pm 0.86}$\\
\textit{ARN+SS (rotation)}  & ${43.90 \pm 0.92}$ & ${57.20 \pm 0.90}$\\
\hline
\multicolumn{3}{c}{\bf Temp./Spat. Self-super. \& Att. by alignment (TSA\! \vs~\!SSA)} \\
\hline
\textit{ARN+TSA (temp. jigsaw)}  & ${45.20 \pm 0.98}$ & ${59.11 \pm 0.86}$\\
\textit{ARN+SSA (spat. jigsaw)}  & ${45.15 \pm 0.96}$ & ${60.56 \pm 0.86}$\\
\textit{ARN+SSA (rotation)}  & ${45.52 \pm 0.96}$ & ${58.96 \pm 0.87}$\\ 

\bottomrule
\end{tabular}}
\end{table*}

\begin{table*}[t]
\centering
\caption{Evaluations on \textit{mini}MIT and UCF101 datasets (given 5-way acc.) See the legend at the top of Table \ref{tab:1} for the description of abbreviations.}
\label{tab:2}
\makebox[\textwidth]{
\setlength{\tabcolsep}{0.4em}
\fontsize{9}{12}\selectfont
\begin{tabular}{lcccc}
\toprule
& \multicolumn{2}{c}{miniMIT} & \multicolumn{2}{c}{UCF101} \\
Model & 1-shot & 5-shot & 1-shot & 5-shot \\ \hline
\textit{C3D Prototypical Net} \cite{snell2017prototypical} & ${33.65 \pm 1.01}$ & ${45.11 \pm 0.90}$ & ${57.05 \pm 1.02}$ & ${78.25 \pm 0.73}$\\
\textit{C3D RelationNet} \cite{sung2017learning} & ${35.71 \pm 1.02}$ & ${47.32 \pm 0.91}$ & ${58.21 \pm 1.02}$ & ${78.35 \pm 0.72}$\\
\textit{C3D SoSN} \cite{zhang2019power} & $40.83 \pm 0.99$ & $52.16 \pm 0.95$ & $62.57 \pm 1.03$ & $81.51 \pm 0.74$\\
\hline
\multicolumn{5}{c}{\bf Temporal/Spatial Attention only (TA \vs~SA)} \\
\hline
\textit{ARN+TA} & ${41.65 \pm 0.97}$ & ${56.75 \pm 0.93}$ & ${63.35 \pm 1.03}$ & ${80.59 \pm 0.77}$\\
\textit{ARN+SA} & ${41.27 \pm 0.98}$ & ${55.69 \pm 0.92}$ & ${63.73  \pm 1.08}$ & ${82.19 \pm 0.70}$\\
\textit{ARN+TA+SA} & ${41.85 \pm 0.99}$ & ${56.43 \pm 0.87}$ & ${64.48 \pm 1.06}$ & ${82.37 \pm 0.72}$\\
\hline
\multicolumn{5}{c}{\bf Temporal/Spatial Self-supervision only (TS \vs~SS)} \\
\hline
\textit{ARN+TS (temp. jigsaw)} &  ${42.45 \pm 0.96}$ & ${54.67 \pm 0.87}$ &  ${63.79 \pm 1.02}$ & ${82.14 \pm 0.77}$\\
\textit{ARN+SS (spat. jigsaw)} & ${42.68 \pm 0.95}$ & ${54.46 \pm 0.88}$ &  ${63.75 \pm 0.98}$ & ${80.92 \pm 0.72}$\\
\textit{ARN+SS (rotation)} & ${42.01 \pm 0.94}$ & ${56.83 \pm 0.86}$ &  ${63.95 \pm 1.03}$ & ${81.09 \pm 0.76}$\\
\hline
\multicolumn{5}{c}{\bf Temp./Spat. Self-super. \& Att. by alignment (TSA\! \vs~\!SSA)} \\
\hline
\textit{ARN+TSA (temp. jigsaw)} & ${42.65 \pm 0.94}$ & ${57.35 \pm 0.85}$ & ${65.46 \pm 1.05}$ & ${82.97 \pm 0.71}$\\
\textit{ARN+SSA (spat. jigsaw)} & ${42.92 \pm 0.95}$ & ${56.21 \pm 0.85}$ & ${66.04 \pm 1.01}$ & ${82.68 \pm 0.72}$\\
\textit{ARN+SSA (rotation)} & ${43.05\pm 0.97}$ & ${56.71 \pm 0.87}$ & ${66.32 \pm 0.99}$ & ${83.12 \pm 0.70}$\\
\bottomrule
\end{tabular}}
\end{table*}

\subsection{Comparison with previous works}
\label{sec:weak}

Section \ref{sec:issue} explains the issues with existing methods, protocols, and the lack of publicly available codes. 
For a fair comparison, we use firstly the protocol of \cite{mishra2018generative} (HMDB51 and UCF101 datasets) but we chose 5 splits at random according to their protocol to average results over multiple runs: we report an average-case result not the best case or a single run result (in contrast to \cite{mishra2018generative}). We also use the Kinetics split of \cite{Zhu_2018_ECCV}, and compare our approach with \cite{mishra2018generative,Zhu_2018_ECCV,dwivedi2019protogan}. 

Table \ref{tab:compar} shows that our ARN (best variant) outperforms GenApp \cite{mishra2018generative}, ProtoGAN \cite{dwivedi2019protogan}, CMN \cite{Zhu_2018_ECCV} by a large margin of 3\% to 10\% on the three protocols.  Our standard errors are low as they result from  5 runs on 5 splits (average case) while a large deviation of ProtoGAN \cite{dwivedi2019protogan} was obtained \wrt~episodes on a single split.

\begin{figure}[t]
    \centering
    \includegraphics[width=\linewidth]{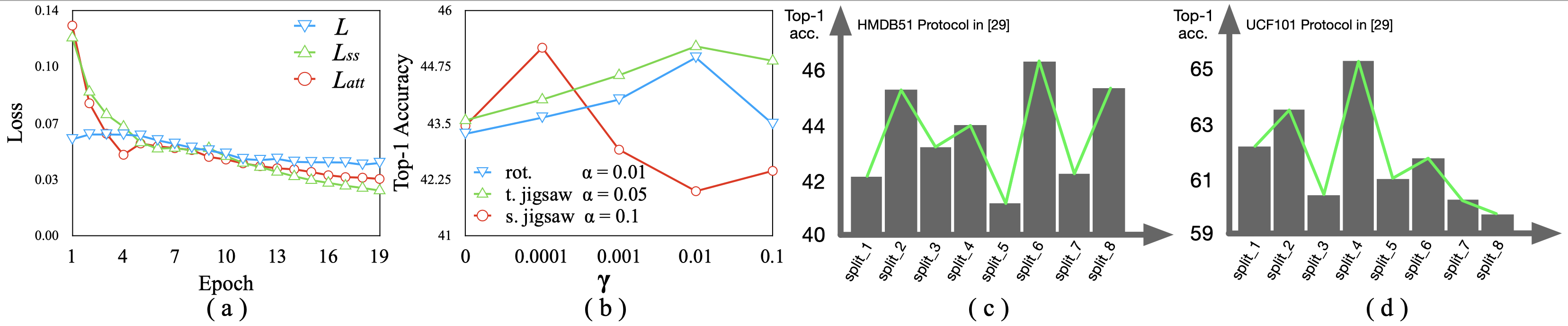}
    \caption{\small In Fig. \ref{fig:ssta_loss}a are the loss curves for Eq. \eqref{eq:finloss}. Fig. \ref{fig:ssta_loss}b shows the validation score \wrt~$\gamma$ (1-shot prot.) Applying Spatial Self-super. \& Attention by alignment (SSA) $\gamma\!>\!0$ outperforms the Spatial Self-super. \& Attention only ($\gamma\!=\!0$). Fig. \ref{fig:ssta_loss}c and \ref{fig:ssta_loss}d show the performance variation on random splits of HMDB51 and UCF101 proposed by \cite{mishra2018generative}.}
    \label{fig:ssta_loss}
\end{figure}

\vspace{0.05cm}
\noindent\textbf{The weakness of  protocol \cite{mishra2018generative}.} Evaluation protocols in \cite{mishra2018generative} rely on randomly selecting training/testing classes with 50--50 ratio from all classes to form  training/testing splits on HMDB51 and UCF101. The performance of that protocol varies heavily due to the randomness. Moreover, results of \cite{mishra2018generative} are reported on a single run. 
Figures \ref{fig:ssta_loss}c and  \ref{fig:ssta_loss}d show up to 6\% variations due to the randomness, making a fair comparison between models difficult on such a protocol. The lack of validation set makes cross-validation also a random process affecting results.

We rectify all this by providing standardized training, validation, and testing splits on HMDB51, \textit{mini}MIT and UCF101.  In what follows, we use our new splits with our few-shot ARN. 
We equip the Prototypical Net \cite{snell2017prototypical}, Relation Net \cite{sung2017learning} and SoSN \cite{zhang2019power} with a 3D conv. feature encoder ({\em C3D}) for baselines used below. 

\vspace{0.05cm}
\noindent\textbf{ARN modules (ablations).}  We start by studying ARN modules on HMDB51. 
Table \ref{tab:att} shows that combining attention with the baseline C3D SoSN pipeline brings $\sim$1\% gain. Switching to the attention by alignment brings 1.2--1.9\% gain over  the naive attention unit. Combining self-supervision with (i) the baseline and (ii) the baseline with attention brings $\sim$3\% and $\sim$3.5\% gain, resp. Combining all units together (attention, self-supervision and alignment) yields $\sim$\textbf{5}\% gain. The  computational cost is 
similar to running either self-supervision or alignment.

Thus, in what follows we will report results for the most distinct four variants: (i) baseline ({\em C3D SoSN}), (ii) Temporal/Spatial Attention only ({\em TA} \& {\em SA}), Temporal/Spatial Self-supervision w/o attention ({\em TS} \& {\em SS}), and Temporal/Spatial Self-supervision with attention by alignment ({\em TSA} \& {\em SSA}).

\begin{table*}[b]
\centering
\caption{\small Comparison of pooling functions on our HMDB51 split (5-way 1-shot).}
\label{tab:pooling}
\makebox[\textwidth]{
\setlength{\tabcolsep}{0.3em}
\fontsize{8}{10}\selectfont
\begin{tabular}{cccccc}
\toprule
No Pooling & Average & Average+PN & Max & Second-order (w/o PN) & Second-order (with PN) \\ \hline
35.71 & 39.51 & 40.02 & 38.95 & 39.97 & $\mathbf{40.83}$ \\
\bottomrule
\end{tabular}}
\end{table*}

\noindent\textbf{Pooling (ablations).} Section \ref{sec:pool} discusses pooling variants from Table \ref{tab:pooling}. Second-order pooling (with PN) outperforms second-order pooling (w/o PN) followed by average and max pooling. Combining average pooling with PN boosts its results which is consistent with the theoretical analysis in Figure \ref{fig:ar_attention}. In what follows, we use the best pooling variant only, that is second-order pooling with PN.

\noindent\textbf{Main evaluations.} Tables \ref{tab:1} and \ref{tab:2} present main evaluations on the proposed by us protocols. Notably, our approaches outperform all baselines (known approaches enhanced by us with the  C3D-based encoder). Below, we detail the results.

\noindent\textbf{Attention.} Tables \ref{tab:1} and  \ref{tab:2} investigate the Temporal and Spatial Attention denoted as ({\em TA}) and  ({\em SA}) on our few-shot ARN. TA on the 1-shot and 5-shot protocols improves the accuracy by $1.0\%$ and $2.5\%$ while SA boosts the 1- and 5-shot accuracy by $0.5\%$ and $1.0\%$, respectively. 
For the combined Temporal and Spatial Attention ({\em TA}+{\em SA}), the  Eq. \eqref{eq:sh} is used with  $\alpha_s\!=\!1.0$ and $\alpha_t\!=\!0.5$ (HMDB51) and $\alpha_s\!=\!1.5$ and $\alpha_t\!=\!1.0$ (UCF101) chosen on the validation split. 
Tables \ref{tab:1} and  \ref{tab:2} show that
SA+TA  achieves a further improvement of up to $1.1\%$ for 1-shot learning but for 5-shot learning it may suffer an $0.8\%$ drop in accuracy compared to the best score of TA and SA while still achieving between an 0.8 and 4.2\% gain over the baseline  C3D SoSN. This  is consistent with our argument that vanilla attention units can be further improved for a better performance. 

\noindent\textbf{Temporal/spatial self-supervision.} In this experiment, we disable attention units. Tables \ref{tab:1} and  \ref{tab:2} show that self-supervision \wrt~either temporal or spatial mode boosts performance of 1-shot and 5-shot learning over the  C3D SoSN baseline on HMDB51 up to 3.2\%. On miniMIT, we observe gains between 1.2 and 4.6\%. On UCF101, we see gains between 0.6 and 1.4\%. However, for UCF101 dataset, self-supervision by the spatial jigsaw and rotation lead to a marginal performance drop on 5-shot learning compared to  C3D SoSN. 

\noindent\textbf{Temporal/spatial self-supervision \& attention by alignment.} 
According to Tables \ref{tab:1} and  \ref{tab:2}, the gains are in 2--5\% range compared to the baseline C3D SoSN.
Figure \ref{fig:ssta_loss}a shows the training loss \wrt~epoch (HMDB51) ({\em temp. jigsaw}). Figure \ref{fig:ssta_loss}b shows the validation accuracy (HMDB51) \wrt~$\gamma$ for SS ({\em rot.}) As can be seen, Self-supervision combined with Attention by alignment (any curve for $\gamma\!>\!0$) scores higher than Self-supervision with Attention only ($\gamma\!=\!0$).

%% file: 5-conclusion.tex
\section{Conclusions}
We have proposed a new few-shot Action Recognition Network (ARN) which comprises an encoder, comparator and an attention mechanism to model short- and long-range temporal patterns. We have investigated the role of self-supervision via spatial and temporal augmentations/auxiliary tasks. Moreover, we have proposed a novel mechanism dubbed {\em attention by alignment} which tackles the so-called distribution shift of temporal positions of discriminative long-range blocks. By combining losses of self-supervision and attention by alignment, we see gains of up to 6\% accuracy. We make our dataset splits publicly available to facilitate fair comparisons of few-shot action recognition pipelines.

%% file: appendix.tex
Below we demonstrate the detailed training/validation/testing splits used in our paper. Though the limited number of previous works propose some evaluation splits on several action recognition datasets, they differ in every paper thus making it very difficult to produce fair comparisons with other works. Additionally, most of works use random train/test splits and have no validation set, thus making the results suffer from high variance and potentially overfitting to the test data. As we aim to fix these problems, we formally introduce three new evaluation protocols as the standard benchmarks, which can help compare models more accurately in a fair setting.

\subsection{HMDB51}
\noindent\textbf{Actions of Train Split (31):} brush hairs, catch, chew, clap, climb, climb stairs, dive, draw sword, dribble, drink, fall floor, flic flac, handstand, hug, jump, kiss, pullup, punch, push, ride bike, ride horse, shake hands, shoot bow, situp, stand, sword, sword exercies, throw, turn, walk, wave. 

\noindent\textbf{Actions of Validation Split (10):} cartwheel, eat, golf, hit, laugh, shoot ball, shoot gun, smile, somersault, swing baseketball.

\noindent\textbf{Actions of Validation Split (10):} fencing, kick, kick ball, pick, pour, pushup, run, sit, smoke, talk.

\subsection{miniMIT}
\noindent\textbf{Actions of Train Split (120):} arresting, assembling, attacking, baking, barbecuing, barking, bending, bicycling, biting, boating, bouncing, brushing, bulldozing, burning, camping, carrying, celebrating, chopping, clapping, cleaning, clinging, closing, combing, competing, covering, crawling, crying, cutting, descending, destroying, digging, dining, drawing, drenching, drilling, drinking, dripping, driving, dropping, drying, dunking, emptying, entering, erupting, falling, filling, flipping, floating, flying, folding, frying, handwriting, hanging, hitting, juggling, kicking, knitting, landing, laughing, leaping, lecturing, lifting, mopping, opening, parading, photographing, picking, placing, pouring, pressing, protesting, pulling, pushing, rafting, raining, reading, removing, repairing, riding, rising, rowing, running, sawing, scratching, sewing, shaking, shaving, shopping, shouting, shredding, singing, skating, sleeping, slicing, sliding, smiling, smoking, snowing, speaking, spraying, spreading, sprinting, stacking, stirring, stitching, stretching, stroking, studying, swimming, swinging, tapping, tattooing, turning, twisting, typing, vacuuming, walking, washing, whistling, wrapping.

\noindent\textbf{Actions of Validation Split (40):} ascending, boiling, bubbling, chasing, combusting, constructing, cracking, crashing, crushing, diving, drumming, eating, exercising, gardening, grilling, grooming, hammering, hugging, inflating, licking, painting, peeling, pitching, planting, playing, playing sports, rolling, sanding, shoveling, smashing, spinning, steering, surfing, sweeping, tapping, throwing, unloading, watering, waving, wrestling.

\noindent\textbf{Actions of Test Split (40):} boxing, carving, catching, cheering, chewing, climbing, colliding, cooking, crafting, dancing, feeding, fishing, flooding, frowning, gripping, hiking, howling, jumping, launching, mowing, overflowing, pedaling, performing, piloting, playing music, racing, raising, resting, rubbing, sailing, slapping, sneezing, sniffing, splashing, storming, tying, waking, waxing, welding, yawning.

\subsection{UCF101}
\noindent\textbf{Actions of Train Split (70):} ApplyEyeMakeUp, Archery, BabyCrawling, BalanceBeam, BandMarching, BaseballPitch, Basketball, BasketballDunk, BenchPress, Biking, Billiards, BlowDryHair, BodyWeightSquats, Bowling, BoxingPunchingBag, BoxingSpeedBag, BreastStroke, BrushingTeeth, CricketBowling, Drumming, Fencing, FieldHockeyPenalty, FrisbeeCatch, FrontCrawl, Haircut, Hammering, HeadMassage, HulaHoop, JavelinThrow, JugglingBalls, JumpingJack, Kayaking, Knitting, LongJump, Lunges, MilitaryParade, Mixing, MoppingFloow, Nunchucks, ParallelBars, PizzaTossing, PlayingCello, PlayingDhol, PlayingFlute, PlayingPiano, PlayingSitar, PlayingTabla, PlayingViolin, PoleVault, Pullups, PushUps, Rafting, RopeClimbing, Rowing, ShavingBeard, Skijet, SoccerJuggling, SoccerPenalty, SumoWrestling, Swing, TableTennisShot, Taichi, ThrowDiscus, TrampolineJumpling, Typing, UnevenBars, WalkingWithDog, WallPushups, WritingOnBoard, YoYo.

\noindent\textbf{Actions of Validation Split (10):} ApplyLipstick, CricketShot, HammerThrow, HandstandPushups, HighJump, HorseRiding, PlayingDaf, PlayingGuitar, Shotput, SkateBoarding.

\noindent\textbf{Actions of Test Split (21):} BlowingCandles, CleanAndJerk, CliffDiving, CuttingInKitchen, Diving, FloorGymnastics, GolfSwing, HandstandWalking, HorseRace, IceDancing, JumpRope, PommelHorse, Punch, RockClimbingIndoor, SalsaSpin, Skiing, SkyDiving, StillRings, Surfing, TennisSwing, VolleyballSpiking.

%% file: 3831.bbl
\begin{thebibliography}{10}
\providecommand{\url}[1]{\texttt{#1}}
\providecommand{\urlprefix}{URL }
\providecommand{\doi}[1]{https://doi.org/#1}

\bibitem{antoniou2018train}
Antoniou, A., Edwards, H., Storkey, A.: How to train your maml. arXiv preprint
  (2018)

\bibitem{BartU05}
Bart, E., Ullman, S.: Cross-generalization: Learning novel classes from a
  single example by feature replacement. In: CVPR (2005)

\bibitem{i3d_net}
Carreira, J., Zisserman, A.: {Quo Vadis, Action Recognition? A New Model and
  the Kinetics Dataset}. In: CVPR (2018)

\bibitem{doersch2015unsupervised}
Doersch, C., Gupta, A., Efros, A.A.: Unsupervised visual representation
  learning by context prediction. In: CVPR (2015)

\bibitem{dosovitskiy2014discriminative}
Dosovitskiy, A., Springenberg, J.T., Riedmiller, M., Brox, T.: Discriminative
  unsupervised feature learning with convolutional neural networks. In: NeurIPS
  (2014)

\bibitem{dwivedi2019protogan}
Dwivedi, S.K., Gupta, V., Mitra, R., Ahmed, S., Jain, A.: Protogan: Towards few
  shot learning for action recognition. arXiv preprint  (2019)

\bibitem{DeepKSPD}
Engin, M., Wang, L., Zhou, L., Liu, X.: Deepkspd: Learning kernel-matrix-based
  {SPD} representation for fine-grained image recognition. In: ECCV. vol.
  11206, pp. 629--645 (2018). \doi{10.1007/978-3-030-01216-8\_38}

\bibitem{fei2006one}
Fei-Fei, L., Fergus, R., Perona, P.: One-shot learning of object categories.
  TPAMI  (2006)

\bibitem{fernando2017self}
Fernando, B., Bilen, H., Gavves, E., Gould, S.: Self-supervised video
  representation learning with odd-one-out networks. In: CVPR (2017)

\bibitem{NIPS2004_2576}
Fink, M.: Object classification from a single example utilizing class relevance
  metrics. In: NeurIPS (2005)

\bibitem{finn2017model}
Finn, C., Abbeel, P., Levine, S.: Model-agnostic meta-learning for fast
  adaptation of deep networks. In: ICML (2017)

\bibitem{gan2018geometry}
Gan, C., Gong, B., Liu, K., Su, H., Guibas, L.J.: Geometry guided convolutional
  neural networks for self-supervised video representation learning. In: CVPR
  (2018)

\bibitem{gnn}
Garcia, V., Bruna, J.: Few-shot learning with graph neural networks. In: ICLR
  (2018)

\bibitem{gidaris2019boosting}
Gidaris, S., Bursuc, A., Komodakis, N., P{\'e}rez, P., Cord, M.: Boosting
  few-shot visual learning with self-supervision. In: ICCV (2019)

\bibitem{Gidaris_2019_CVPR}
Gidaris, S., Komodakis, N.: Generating classification weights with gnn
  denoising autoencoders for few-shot learning. In: CVPR (2019)

\bibitem{gidaris2018unsupervised}
Gidaris, S., Singh, P., Komodakis, N.: Unsupervised representation learning by
  predicting image rotations. arXiv preprint  (2018)

\bibitem{guo2018neural}
Guo, M., Chou, E., Huang, D.A., Song, S., Yeung, S., Fei-Fei, L.: Neural graph
  matching networks for fewshot 3d action recognition. In: ECCV (2018)

\bibitem{hariharan2017low}
Hariharan, B., Girshick, R.: Low-shot visual recognition by shrinking and
  hallucinating features. In: ICCV (2017)

\bibitem{jian2020representation}
Jian, S., Hu, L., Cao, L., Lu, K.: Representation learning with multiple
  lipschitz-constrained alignments on partially-labeled cross-domain data. In:
  AAAI. pp. 4320--4327 (2020)

\bibitem{jian2019evolutionarily}
Jian, S., Hu, L., Cao, L., Lu, K., Gao, H.: Evolutionarily learning
  multi-aspect interactions and influences from network structure and node
  content. In: Proceedings of the AAAI Conference on Artificial Intelligence.
  vol.~33, pp. 598--605 (2019)

\bibitem{Kim_2019_CVPR}
Kim, J., Kim, T., Kim, S., Yoo, C.D.: Edge-labeling graph neural network for
  few-shot learning. In: CVPR (2019)

\bibitem{koch2015siamese}
Koch, G., Zemel, R., Salakhutdinov, R.: Siamese neural networks for one-shot
  image recognition. In: ICML Deep Learning Workshop (2015)

\bibitem{koniusz2016tensor}
Koniusz, P., Cherian, A., Porikli, F.: Tensor representations via kernel
  linearization for action recognition from 3d skeletons. In: ECCV (2016)

\bibitem{koniusz2017domain}
Koniusz, P., Tas, Y., Porikli, F.: Domain adaptation by mixture of alignments
  of second-or higher-order scatter tensors. In: CVPR (2017)

\bibitem{pk_tpami_ar}
Koniusz, P., Wang, L., Cherian, A.: Tensor representations for action
  recognition. TPAMI  (2020)

\bibitem{koniusz2017higher}
Koniusz, P., Yan, F., Gosselin, P.H., Mikolajczyk, K.: Higher-order occurrence
  pooling for bags-of-words: Visual concept detection. TPAMI  (2017)

\bibitem{pk_hz_tpami_fsl}
Koniusz, P., Zhang, H.: Power normalizations in fine-grained image, few-shot
  image and graph classification. TPAMI  (2020)

\bibitem{koniusz2018deeper}
Koniusz, P., Zhang, H., Porikli, F.: A deeper look at power normalizations. In:
  CVPR (2018)

\bibitem{Kuehne11}
Kuehne, H., Jhuang, H., Garrote, E., Poggio, T., Serre, T.: {HMDB}: a large
  video database for human motion recognition. In: ICCV (2011)

\bibitem{lai2020mast}
Lai, Z., Lu, E., Xie, W.: Mast: A memory-augmented self-supervised tracker. In:
  CVPR (2020)

\bibitem{lake_oneshot}
Lake, B.M., Salakhutdinov, R., Gross, J., Tenenbaum, J.B.: One shot learning of
  simple visual concepts. CogSci  (2011)

\bibitem{lee2019meta}
Lee, K., Maji, S., Ravichandran, A., Soatto, S.: Meta-learning with
  differentiable convex optimization. In: CVPR (2019)

\bibitem{Li9596}
Li, F.F., VanRullen, R., Koch, C., Perona, P.: Rapid natural scene
  categorization in the near absence of attention. Proceedings of the National
  Academy of Sciences  (2002)

\bibitem{miller_one_example}
Miller, E.G., Matsakis, N.E., Viola, P.A.: Learning from one example through
  shared densities on transforms. In: CVPR (2000)

\bibitem{mishra2018generative}
Mishra, A., Verma, V.K., Reddy, M.S.K., Arulkumar, S., Rai, P., Mittal, A.: A
  generative approach to zero-shot and few-shot action recognition. In: WACV
  (2018)

\bibitem{monfortmoments}
Monfort, M., Andonian, A., Zhou, B., Ramakrishnan, K., Bargal, S.A., Yan, T.,
  Brown, L., Fan, Q., Gutfruend, D., Vondrick, C., Oliva, A.: Moments in time
  dataset: one million videos for event understanding. TPAMI  (2019)

\bibitem{elbcm_brod}
Romero, A., Ter{\'{a}}n, M.Y., Gouiff{\`{e}}s, M., Lacassagne, L.: Enhanced
  local binary covariance matrices {(ELBCM)} for texture analysis and object
  tracking. MIRAGE  (2013)

\bibitem{ILSVRC15}
Russakovsky, O., Deng, J., Su, H., Krause, J., Satheesh, S., Ma, S., Huang, Z.,
  Karpathy, A., Khosla, A., Bernstein, M., Berg, A.C., Fei-Fei, L.: {ImageNet}
  large scale visual recognition challenge. IJCV  (2015)

\bibitem{Rusu2019LEO}
Rusu, A.A., Rao, D., Sygnowski, J., Vinyals, O., Pascanu, R., Osindero, S.,
  Hadsell, R.: Meta-learning with latent embedding optimization. In: ICLR
  (2019)

\bibitem{sermanet2017time}
Sermanet, P., Lynch, C., Chebotar, Y., Hsu, J., Jang, E., Schaal, S., Levine,
  S.: Time-contrastive networks: Self-supervised learning from pixels. In: ICRA
  (2017)

\bibitem{Simon_2019_NIPS}
Simon, C., Koniusz, P., Nock, R., Harandi, M.: Deep subspace networks for
  few-shot learning. In: NeurIPS workshops (2019)

\bibitem{Simon_2020_CVPR}
Simon, C., Koniusz, P., Nock, R., Harandi, M.: Adaptive subspaces for few-shot
  learning. In: CVPR (2020)

\bibitem{Simon_2020_ECCV}
Simon, C., Koniusz, P., Nock, R., Harandi, M.: On modulating the gradient for
  meta-learning. In: ECCV (2020)

\bibitem{snell2017prototypical}
Snell, J., Swersky, K., Zemel, R.: Prototypical networks for few-shot learning.
  In: NeurIPS (2017)

\bibitem{soomro2012ucf101}
Soomro, K., Zamir, A.R., Shah, M.: Ucf101: A dataset of 101 human actions
  classes from videos in the wild. arXiv preprint  (2012)

\bibitem{su2019boosting}
Su, J.C., Maji, S., Hariharan, B.: Boosting supervision with self-supervision
  for few-shot learning. arXiv preprint  (2019)

\bibitem{sung2017learning}
Sung, F., Yang, Y., Zhang, L., Xiang, T., Torr, P.H., Hospedales, T.M.:
  Learning to compare: Relation network for few-shot learning. In: CVPR (2018)

\bibitem{tuzel_rc}
Tuzel, O., Porikli, F., Meer, P.: Region covariance: {A} fast descriptor for
  detection and classification. In: ECCV (2006)

\bibitem{vinyals2016matching}
Vinyals, O., Blundell, C., Lillicrap, T., Wierstra, D., et~al.: Matching
  networks for one shot learning. In: NeurIPS (2016)

\bibitem{lei_beycov}
Wang, L., Zhang, J., Zhou, L., Tang, C., Li, W.: Beyond covariance: Feature
  representation with nonlinear kernel matrices. In: ICCV. pp. 4570--4578
  (2015). \doi{10.1109/ICCV.2015.519}

\bibitem{wertheimer2019few}
Wertheimer, D., Hariharan, B.: Few-shot learning with localization in realistic
  settings. In: CVPR (2019)

\bibitem{xu2018dense}
Xu, B., Ye, H., Zheng, Y., Wang, H., Luwang, T., Jiang, Y.G.: Dense dilated
  network for few shot action recognition. In: ICMR (2018)

\bibitem{Xu_2019_CVPR}
Xu, D., Xiao, J., Zhao, Z., Shao, J., Xie, D., Zhuang, Y.: Self-supervised
  spatiotemporal learning via video clip order prediction. In: CVPR (2019)

\bibitem{zhang2019power}
Zhang, H., Koniusz, P.: Power normalizing second-order similarity network for
  few-shot learning. In: WACV (2019)

\bibitem{zhang2019few}
Zhang, H., Zhang, J., Koniusz, P.: Few-shot learning via saliency-guided
  hallucination of samples. In: CVPR (2019)

\bibitem{places_dataset}
Zhou, B., Lapedriza, A., Xiao, J., Torralba, A., Oliva, A.: Learning deep
  features for scene recognition using places database. NeurIPS  (2014)

\bibitem{zhu2020self}
Zhu, F., Zhang, L., Fu, Y., Guo, G., Xie, W.: Self-supervised video object
  segmentation. arXiv preprint  (2020)

\bibitem{Zhu_2018_ECCV}
Zhu, L., Yang, Y.: Compound memory networks for few-shot video classification.
  In: ECCV (2018)

\bibitem{Zintgraf2019Cavia}
Zintgraf, L., Shiarli, K., Kurin, V., Hofmann, K., Whiteson, S.: Fast context
  adaptation via meta-learning. In: ICML (2019)

\end{thebibliography}
